\documentclass[11pt]{article}

\newif\ifneuripsstyle
\IfFileExists{neurips_2026.sty}{%
  \neuripsstyletrue
  \PassOptionsToPackage{numbers}{natbib}
  \usepackage[preprint]{neurips_2026}
}{%
  \neuripsstylefalse
  \usepackage[margin=0.82in]{geometry}
  \usepackage{newtxtext}
  \setlength{\parskip}{4pt}
  \setlength{\parindent}{0pt}
}

\usepackage{amsmath,amssymb,amsthm,mathtools}
\usepackage{booktabs}
\usepackage{microtype}
\usepackage{graphicx}
\usepackage{enumitem}
\usepackage{hyperref}
\hypersetup{hidelinks}
\usepackage{array}
\usepackage{algorithm}
\usepackage{float}
\usepackage{algpseudocode}
\usepackage{multirow}
\usepackage{xcolor}
\AtBeginDocument{%
  \DeclareFontShape{OT1}{ptm}{m}{scit}{<->ssub*ptm/m/sc}{}%
  \DeclareFontShape{OT1}{ptm}{b}{scit}{<->ssub*ptm/bx/sc}{}%
}

\ifneuripsstyle
\workshoptitle{Human-AI Coevolution: Measuring Human-Agent Teams in the Agentic Era}
\fi

\title{Beyond ``AI Helps Humans'': Decision-Targeted Evaluation Design for Human--Agent Teams in the Agentic Era}
\author{%
  Hamed Khosravi \\
  H. Milton Stewart School of Industrial and Systems Engineering \\
  Georgia Institute of Technology \\
  Atlanta, GA 30332 \\
  \texttt{hkhosravi7@gatech.edu} \\
  \And
  Xiaoming Huo \\
  H. Milton Stewart School of Industrial and Systems Engineering \\
  Georgia Institute of Technology \\
  Atlanta, GA 30332 \\
  \texttt{huo@gatech.edu}
}
\ifneuripsstyle\else\date{}\fi

\newtheorem{theorem}{Theorem}[section]
\newtheorem{proposition}[theorem]{Proposition}

\newcommand{\E}{\mathbb{E}}
\newcommand{\Var}{\operatorname{Var}}
\newcommand{\Ep}{\widetilde{\mathbb{E}}_{\mathrm{p}}}
\newcommand{\Varp}{\widetilde{\operatorname{Var}}_{\mathrm{p}}}

\newcommand{\TEAM}{\textsc{TEAM-Design}}

\newcommand{\RBotTeam}{\resultblank}
\newcommand{\RBotNeyman}{\resultblank}

\IfFileExists{tables/results_macros.tex}{

\providecommand{\RBotNeyman}{}\renewcommand{\RBotNeyman}{65.1}

\providecommand{\RBotTeam}{}\renewcommand{\RBotTeam}{77.1}

\providecommand{\RDeltaBotNeyman}{}\renewcommand{\RDeltaBotNeyman}{25.4}

\providecommand{\RDeltaBotTeam}{}\renewcommand{\RDeltaBotTeam}{30.9}

\providecommand{\RDeltaMin}{}\renewcommand{\RDeltaMin}{0.02}

\providecommand{\RFairBatchPow}{}\renewcommand{\RFairBatchPow}{75.2}

\providecommand{\RFairEIGPow}{}\renewcommand{\RFairEIGPow}{59.5}

\providecommand{\RFairEqualPow}{}\renewcommand{\RFairEqualPow}{60.1}

\providecommand{\RFairTeamPow}{}\renewcommand{\RFairTeamPow}{76.3}

\providecommand{\RGainDelta}{}\renewcommand{\RGainDelta}{5.5}
\providecommand{\RGainDeltaZ}{}\renewcommand{\RGainDeltaZ}{14.2}

\providecommand{\RPowBalEIG}{}\renewcommand{\RPowBalEIG}{40,000}

\providecommand{\RPowBalTeam}{}\renewcommand{\RPowBalTeam}{52,000}
\providecommand{\RPowBalTeamB}{}\renewcommand{\RPowBalTeamB}{52,000}
\providecommand{\RPowEasyEIGB}{}\renewcommand{\RPowEasyEIGB}{6,000}
\providecommand{\RPowEasyTeamB}{}\renewcommand{\RPowEasyTeamB}{8,000}
\providecommand{\RPowGapEIG}{}\renewcommand{\RPowGapEIG}{26,000}
\providecommand{\RPowGapEIGB}{}\renewcommand{\RPowGapEIGB}{40,000}
\providecommand{\RPowGapTeam}{}\renewcommand{\RPowGapTeam}{20,000}
\providecommand{\RPowGapTeamB}{}\renewcommand{\RPowGapTeamB}{26,000}

\providecommand{\RPriorDiffuse}{}\renewcommand{\RPriorDiffuse}{77.6}
\providecommand{\RPriorMisspec}{}\renewcommand{\RPriorMisspec}{77.6}
\providecommand{\RPriorWell}{}\renewcommand{\RPriorWell}{77.9}

\providecommand{\RSurrMmxPow}{}\renewcommand{\RSurrMmxPow}{70.2}
\providecommand{\RSurrMmxSplit}{}\renewcommand{\RSurrMmxSplit}{0.962}
\providecommand{\RSurrOptPow}{}\renewcommand{\RSurrOptPow}{81.3}
\providecommand{\RSurrOptSplit}{}\renewcommand{\RSurrOptSplit}{0.907}
\providecommand{\RSurrStabPow}{}\renewcommand{\RSurrStabPow}{81.1}
\providecommand{\RSurrStabSplit}{}\renewcommand{\RSurrStabSplit}{0.893}
\providecommand{\RTauBest}{}\renewcommand{\RTauBest}{0.03}
\providecommand{\RTauBestPow}{}\renewcommand{\RTauBestPow}{75.3}
\providecommand{\RTauHi}{}\renewcommand{\RTauHi}{0.1}
\providecommand{\RTauHiPow}{}\renewcommand{\RTauHiPow}{70.0}
\providecommand{\RTauLo}{}\renewcommand{\RTauLo}{0.0025}
\providecommand{\RTauLoPow}{}\renewcommand{\RTauLoPow}{59.5}
\providecommand{\RXoverHiEIG}{}\renewcommand{\RXoverHiEIG}{58.8}
\providecommand{\RXoverHiTeam}{}\renewcommand{\RXoverHiTeam}{76.6}
\providecommand{\RXoverLoEIG}{}\renewcommand{\RXoverLoEIG}{31.3}
\providecommand{\RXoverLoTeam}{}\renewcommand{\RXoverLoTeam}{32.9}
\providecommand{\RXoverN}{}\renewcommand{\RXoverN}{8,000}

\providecommand{\SFairBatch}{}\renewcommand{\SFairBatch}{0.77}
\providecommand{\SFairEIG}{}\renewcommand{\SFairEIG}{0.50}

\providecommand{\SFairTeam}{}\renewcommand{\SFairTeam}{0.83}

}{}
\IfFileExists{tables/clinical_macros.tex}{
\providecommand{\RClinBeatsBoth}{}\renewcommand{\RClinBeatsBoth}{0}
\providecommand{\RClinCells}{}\renewcommand{\RClinCells}{6}

\providecommand{\RClinReads}{}\renewcommand{\RClinReads}{8,723}

}{}
\IfFileExists{tables/semisynth_macros.tex}{

\providecommand{\RSSFullDec}{}\renewcommand{\RSSFullDec}{9.0}

\providecommand{\RSSLoNeyman}{}\renewcommand{\RSSLoNeyman}{15.7}
\providecommand{\RSSLoTeam}{}\renewcommand{\RSSLoTeam}{10.4}

\providecommand{\RSSReps}{}\renewcommand{\RSSReps}{200}

\providecommand{\RSSThresh}{}\renewcommand{\RSSThresh}{16}
\providecommand{\RSSTruthA}{}\renewcommand{\RSSTruthA}{10}

}{}
\IfFileExists{tables/floor_macros.tex}{
\providecommand{\RFloorCells}{}\renewcommand{\RFloorCells}{36}
\providecommand{\RFloorDetHi}{}\renewcommand{\RFloorDetHi}{100}
\providecommand{\RFloorDetLo}{}\renewcommand{\RFloorDetLo}{77.5}
\providecommand{\RFloorDetMid}{}\renewcommand{\RFloorDetMid}{98.0}
\providecommand{\RFloorLead}{}\renewcommand{\RFloorLead}{10}

\providecommand{\RFloorP}{}\renewcommand{\RFloorP}{0.008}
\providecommand{\RFloorShare}{}\renewcommand{\RFloorShare}{0.40}
\providecommand{\RFloorShipped}{}\renewcommand{\RFloorShipped}{0.05}

\providecommand{\RFloorTrail}{}\renewcommand{\RFloorTrail}{25}
\providecommand{\RFloorUntied}{}\renewcommand{\RFloorUntied}{35}
}{}

\begin{document}
\maketitle

\begin{abstract}
Wherever a coding agent works under engineer supervision, or a clinical model assists a radiologist, the deployment question is whether to keep the human--AI workflow or replace it with the human alone or the agent alone. The human--AI workflow is worth keeping only if it beats both of those alternatives. Yet once it is deployed, neither alternative outcome is observed: recovering one means replaying the task with the human alone or with the agent alone, and every replay costs expert time or compute. Under a fixed replay budget, the design question is therefore which tasks should be more likely to receive a human-only replay, and which an agent-only replay. Existing methods do not directly target this decision. Agent benchmarks do not choose which missing baseline to measure, variance-based sampling ignores which of the two comparisons is closer to failing, and Bayesian information methods focus on learning model parameters instead of making the deployment decision. We propose \TEAM, a rule that gives every task two replay probabilities, one per baseline. It raises a probability where the missing baseline outcome is hard to predict from what is already known about the task and where that comparison is harder to establish, and lowers it where replay is expensive. We prove that the rule solves this budgeted design problem, and that drawing the replays at random from recorded probabilities still controls the chance of wrongly declaring that the human--AI workflow beats both. We reanalyze \RClinCells\ clinical settings, where no human--AI workflow beats both alternatives, and a coding benchmark, where one does, then evaluate \TEAM\ on synthetic designs and on a semi-synthetic design built from a real chest X-ray reader study. \TEAM\ works best when one of the two comparisons is clearly harder to settle than the other, and can do worse than variance-based allocation when the two are similarly difficult.
\end{abstract}

\section{Introduction}
\label{sec:intro}

\paragraph{The deployment question.}
Human and AI systems increasingly work together. In software, coding agents write code, use tools, and run tests while engineers review and redirect their work. Anthropic reports that its employees use Claude frequently yet can fully delegate only $0$--$20\%$ of their work \cite{anthropicwork}, and a fintech team deploying Claude Code is moving toward developers reviewing the agent's pull requests \cite{credcase}. In medicine, AI models assist clinicians who remain responsible for the final decision. In both settings, the organization observes the outcome of the human--AI workflow it actually runs, in which the human and the agent work the same task together. What it does not observe is how the same task would have turned out with the human alone or the agent alone. Hence the deployment question we study: is the human--AI workflow worth keeping if either alternative could be used instead?

\paragraph{The missing baselines.}
Answering this requires knowing how the same task would have turned out under each alternative, but neither outcome is recorded. An agent-only replay uses model calls, compute, and tools. A human-only replay may require a senior engineer to redo the task independently, or a clinician to read the case without AI assistance. Running both alternatives on every task can cost more than the evaluation itself is worth, which leaves a practical problem.
\begin{quote}
\emph{Given a fixed replay budget, which tasks should be more likely to receive a human-only replay, and which an agent-only replay, so that we can determine whether the human--AI workflow beats both alternatives?}
\end{quote}
We do not choose a fixed subset of tasks in advance. Each task receives two replay probabilities, one per baseline, and the two replay decisions are drawn separately, so a task may receive neither replay, one, or both. A higher probability means that baseline is replayed more often among tasks carrying similar information.

\paragraph{The two settings.}
In coding, human and agent replays differ greatly in cost. In medicine, some studies have clinicians evaluate the same cases with and without AI support \cite{collabcxr,chanda2024}, so we use the clinical data to study the two comparisons and coding to motivate unequal replay costs. Some coding benchmarks report one overall score for each of the human alone, the agent alone, and the human--AI workflow \cite{haieval2025,swebench2024}. Those are averages across tasks, not three outcomes recorded on each individual task, so no task has a gain to estimate and selective replay cannot be tested on them.

\paragraph{Related work.}
Prior work provides pieces of this problem, but not the full evaluation design. A framework already compares the human alone, human with AI, and AI alone \cite{benmichael2025}, but it does not decide which tasks should receive a human-only or agent-only replay. Work on agent evaluation shows that accuracy alone can miss evaluation cost and deployment validity \cite{kapoor2024,reuel2024}. Active learning and experimental design choose data for a downstream decision \cite{filstroff2024,rossa2026}, and off-policy methods design collection for estimating policy performance \cite{wang2017,douglas2026}. Classical variance-based methods, including Neyman allocation and cost-aware sampling, spend measurements where outcomes are uncertain and hold back where measurement is costly \cite{angelopoulos2025cost,gilbert2013}, but they do not account for which of our two comparisons is closer to failing. Ranking and best-system selection methods also allocate measurements efficiently \cite{bechhofer1954,chen2000,kimnelson2006,audibert2010}, but they typically choose among outcomes that can be measured directly. Sizing a second stage from a first-stage estimate is classical two-stage sampling \cite{stein1945}. Here, both baseline outcomes are missing and costly to obtain, and the human--AI workflow must beat both of them.

\paragraph{Contributions.}
We make four contributions.

\begin{itemize}[leftmargin=*,itemsep=2pt,topsep=4pt,parsep=0pt]
\item \textbf{A design that targets the deployment decision.} We give the first evaluation design that chooses which missing baseline to measure on which task, so that a fixed replay budget goes as far as possible toward settling whether the human--AI workflow beats both alternatives. \TEAM\ assigns every task a human-only and an agent-only replay probability in closed form, buying more replay where the missing baseline is hard to predict and where that comparison is harder to establish, and less where replay is expensive. The quantities it needs are learned once from a small \emph{pilot} set of tasks on which the workflow, the human alone, and the agent alone have all been run.

\item \textbf{The rule is optimal, not heuristic.} We show that this design problem is a convex minimax program: because the workflow must beat both alternatives, the objective maximizes over the two comparisons to select whichever needs more evidence, then minimizes over the replay probabilities to make that one as easy as possible to resolve. We prove that the square-root rule of \TEAM\ is its exact solution.

\item \textbf{The decision stays valid.} We prove that choosing the replay probabilities from the pilot and then drawing the replays at random from the recorded probabilities still controls the chance of wrongly declaring that the workflow beats both, and that requiring both bounds to clear needs no multiplicity correction.

\item \textbf{The problem is real, and the rule works.} Across \RClinCells\ clinical study settings, assisted clinicians perform better than clinicians alone but worse than the model alone, so none of these human--AI workflows beats both alternatives. A coding benchmark designed for collaboration shows the opposite pattern. On a semi-synthetic design built from a real chest X-ray study, \TEAM\ gives the lowest estimation error for the gain against the harder baseline at every budget below full replay. Its advantage is strongest when one comparison is clearly harder than the other, and it can perform worse when the two are similarly difficult.
\end{itemize}

\paragraph{Paper organization.}
Section~\ref{sec:problem} defines the evaluation problem. Section~\ref{sec:team-design} develops \TEAM. Section~\ref{sec:theory} gives the theoretical guarantees. Section~\ref{sec:results} evaluates the method and studies when it helps.

\section{Problem formulation}
\label{sec:problem}

A task can be handled three ways: by the human and the agent together, by the human alone, or by the agent alone. Only the first is deployed, and \emph{the workflow} always refers to that joint configuration, never to a human-only or agent-only one; the other two are its \emph{baselines}. For a task from the deployment population we observe information $W\in\mathcal W$, such as tool calls, test failures, or runtime, together with the workflow's outcome $T$. The baseline outcomes $Y_H$ and $Y_A$ are missing unless the task is replayed.

The workflow's average gain over each baseline is
\begin{equation}
\Delta_H=\E[T-Y_H],
\qquad
\Delta_A=\E[T-Y_A],
\label{eq:margins}
\end{equation}
where $\E[\cdot]$ averages over tasks. We write $j\in\{H,A\}$ for either baseline, so $\Delta_j>0$ means the workflow beats baseline $j$ on average. Let $\delta_j\geq0$ be the improvement required over that baseline. The workflow is kept only when
\begin{equation}
\Delta_H>\delta_H
\quad\text{and}\quad
\Delta_A>\delta_A;
\label{eq:both}
\end{equation}
setting $\delta_H=\delta_A=0$ asks only whether it beats both alternatives.

Now take an evaluation pool of $n$ independent tasks, subscripted by $i$. We observe $(W_i,T_i)$, while $Y_{ij}$ is missing unless baseline $j$ is replayed at cost $c_j(W_i)$. Since we cannot replay both baselines on every task, we choose a pair of \emph{replay rules}. A rule is a function of the task information, $q_j:\mathcal W\to[q_{j,\min},1]$, so a task with information $W$ has baseline $j$ replayed with probability $q_j(W)$, and $R_j\mid W\sim\mathrm{Bernoulli}\{q_j(W)\}$ indicates that it happens. The two draws are separate, so a task may receive neither replay, one, or both.

Let $B$ be the average replay budget per task, so that
\begin{equation}
\E\!\left[
c_H(W)q_H(W)+c_A(W)q_A(W)
\right]\leq B,
\qquad
q_j(W)\in[q_{j,\min},1].
\label{eq:budget}
\end{equation}
The floor $q_{j,\min}>0$ leaves every task some chance of revealing each baseline.

Write $\sigma_j^2(q_j)/n$ for the approximate variance of the gain estimate built from the replayed tasks. Replay is what makes this variance a design choice: baseline $j$ is observed only on the tasks where it is replayed, so raising $q_j(W)$ supplies more of the outcomes that inform comparison $j$ and lowers $\sigma_j^2$. The variance therefore depends on the rule as a whole and not on any single probability, and \eqref{eq:variance} gives the exact form. A comparison is hard when that estimate is noisy or when the true gain sits close to the improvement required of it: for $\Delta_j>\delta_j$, the number of tasks needed to establish comparison $j$ grows with $\sigma_j^2(q_j)/(\Delta_j-\delta_j)^2$, which we take as its difficulty. Both comparisons must pass, so improving an already easy one does not resolve the decision. The harder comparison therefore sets the evidence required, and we choose both replay rules to make it as easy as possible:
\begin{equation}
\min_{q_H(\cdot),\,q_A(\cdot)}
\max_{j\in\{H,A\}}
\frac{\sigma_j^2(q_j)}
     {(\Delta_j-\delta_j)^2}
\qquad
\text{subject to \eqref{eq:budget}.}
\label{eq:minimax}
\end{equation}
The variables in \eqref{eq:minimax} are the two rules themselves, one probability for every value of the task information, and \eqref{eq:budget} defines the feasible set they are drawn from. Comparison $j$ depends only on its own rule $q_j$; the two are coupled through the shared budget \eqref{eq:budget} alone, so replay spent on one comparison is replay the other cannot have. Both $\Delta_j$ and $\sigma_j^2(q_j)$ depend on the missing baseline outcomes, so neither is known in advance. Section~\ref{sec:team-design} introduces \TEAM, which estimates them from a pilot and uses them to set the replay probabilities.

\section{\TEAM}
\label{sec:team-design}

Section~\ref{sec:problem} asks for two replay rules, $q_H$ and $q_A$. The best rules depend on quantities we do not know: how close each comparison sits to its threshold, and where the missing baseline outcomes are hard to predict. \TEAM\ estimates both from a small separate pilot, then uses them to set the rules for the evaluation pool.

\paragraph{Learning from a pilot.}
A \emph{pilot} is a set of $n_{\mathrm{p}}$ tasks, separate from the evaluation pool, on which all three configurations are run, so we observe $T$, $Y_H$, and $Y_A$ together; write $\Ep$ for an average over them. Decoration carries the source from here on: a tilde marks a quantity computed on the pilot and used to plan the replay, a hat marks one computed on the evaluation pool and reported, and a plain symbol is the population quantity they estimate. Running all three configurations is expensive, so the pilot is small, and its cost sits outside the budget $B$ of \eqref{eq:budget}, which governs the evaluation pool.

With both baselines observed, the pilot estimates the gain directly as $\widetilde\Delta_j=\Ep[T-Y_j]$, which is \eqref{eq:margins} with the pilot average in place of the deployment average. \TEAM\ uses $\widetilde\Delta_j$ only to plan the replay; the reported answer comes from the evaluation pool.

The pilot also shows which missing outcomes are hard to predict from $W$. Let $\widetilde m_j$ be a predictor of the baseline outcome $Y_j$ fitted on the pilot, and write
\begin{equation}
s_j(W)=\E\!\left[\{Y_j-\widetilde m_j(W)\}^2\;\middle|\;W\right]
\label{eq:sdiff}
\end{equation}
for its expected squared error on tasks with information $W$, the \emph{prediction difficulty} of baseline $j$ there. Conditionally on the pilot, $\widetilde m_j$ is a fixed function, so $s_j$ is a population quantity. A larger $s_j(W)$ means the baseline is harder to predict, so replaying it reveals more. Write $\widetilde s_j(W)$ for the pilot estimate. Appendix~\ref{app:splits} gives both models and the held-out fitting that keeps $\widetilde s_j$ from making a baseline look more predictable than it is.

\paragraph{Estimating the gains after selective replay.}
Take the $n$ evaluation tasks, disjoint from the pilot and indexed by $i=1,\ldots,n$. We observe $(W_i,T_i)$, and $Y_{ij}$ only when baseline $j$ is replayed, which happens with probability $q_j(W_i)$ and is recorded by $R_{ij}\in\{0,1\}$. For each task, define a score $\psi_{ij}$ that estimates the gain over baseline $j$ from that task alone,
\begin{equation}
\psi_{ij}
=
T_i-\widetilde m_j(W_i)
-
\frac{R_{ij}}{q_j(W_i)}
\left\{
Y_{ij}-\widetilde m_j(W_i)
\right\}.
\label{eq:score}
\end{equation}
The first term uses the predicted baseline in place of the missing one; when the task is replayed, the second corrects that prediction by the observed error. The factor $1/q_j(W_i)$ accounts for how rarely corrections are seen: at $q_j(W_i)=0.2$ only one task in five reveals $Y_{ij}$, so an observed correction carries weight $5$. When $R_{ij}=0$ the correction vanishes and the missing $Y_{ij}$ is never needed.

Averaging these scores gives the estimated gain over baseline $j$,
\begin{equation}
\widehat\Delta_j
=
\frac{1}{n}
\sum_{i=1}^{n}
\psi_{ij}.
\label{eq:margin-estimator}
\end{equation}
Because the replay probability is known, $\E[R_{ij}/q_j(W_i)\mid W_i]=1$ and $\E[\widehat\Delta_j]=\Delta_j$ however poorly the pilot predicts; a poor predictor only makes the estimate noisier.

\paragraph{Choosing the replay probabilities.}
The rule needs three things: how much variance replay can remove, how close each comparison sits to its threshold, and what a replay costs. We take the first two in turn, then put them together.

Write $\sigma_j^2(q_j)$ for the variance of one task's score under rule $q_j$, so that $\widehat\Delta_j$ has variance $\sigma_j^2(q_j)/n$. It splits into a part replay cannot touch and a part it can:
\begin{equation}
\sigma_j^2(q_j)
=
\kappa_j
+
\E\!\left[
\frac{s_j(W)}{q_j(W)}
\right].
\label{eq:variance}
\end{equation}
Appendix~\ref{app:variance} derives this. The correction in \eqref{eq:score} has mean zero given the task, because replay is drawn from $W$ alone, so it does not covary with $T-Y_j$; its own variance carries a factor $\{1-q_j(W)\}/q_j(W)$. The first term, $\kappa_j=\Var(T-Y_j)-\E[s_j(W)]$ with $\Var$ the population variance, holds no $q_j$ and is therefore fixed whatever we buy; $\widetilde\kappa_j$ denotes its pilot estimate, formed by the same expression with pilot averages. In the second, a task with information $W$ contributes its prediction difficulty $s_j(W)$ of \eqref{eq:sdiff} divided by its replay probability, so replay buys the most where the missing baseline is hardest to predict. Setting $q_j\equiv1$ returns $\sigma_j^2=\Var(T-Y_j)$, the variance when no outcome is missing, which is what the $-\E[s_j(W)]$ in $\kappa_j$ is for.

The design also needs to know how close comparison $j$ sits to its threshold $\delta_j$. The true gap is unknown, so we use the pilot estimate, held away from zero to stop a noisy estimate that lands near its threshold from taking almost the entire budget:
\begin{equation}
\widetilde d_j^2
=
\left\{
\max(\widetilde\Delta_j-\delta_j,0)
\right\}^2
+
\tau^2,
\label{eq:stabilized-gap}
\end{equation}
where $\tau>0$ is a small constant fixed in advance and the maximum treats the gap as zero when the pilot gain falls below its threshold. This $\widetilde d_j^2$ stands in for the unknown $(\Delta_j-\delta_j)^2$ of Section~\ref{sec:problem}, and Section~\ref{sec:scope} reports that $\tau$ works best on the scale of the pilot's uncertainty.

Difficulty is variance over squared gap, so the estimated difficulty of comparison $j$ is \eqref{eq:variance} divided by $\widetilde d_j^2$, with $\widetilde\kappa_j$ and $\widetilde s_j$ in place of the population quantities. Two multipliers turn the constrained problem \eqref{eq:minimax} into one that can be solved task by task: $\lambda_H,\lambda_A\ge0$ with $\lambda_H+\lambda_A=1$ price the two comparisons, and $\eta>0$ prices the budget. A task with information $W$ then contributes
\begin{equation*}
\frac{\lambda_j}{\widetilde d_j^2}\,\frac{\widetilde s_j(W)}{q_j(W)}
\;+\;
\eta\,c_j(W)\,q_j(W),
\end{equation*}
difficulty on the left and cost on the right. The first term falls like $1/q_j(W)$ and the second grows in proportion to $q_j(W)$, so the sum is smallest where the two balance, at
\begin{equation}
q_j^\star(W)
=
\operatorname{clip}_{[q_{j,\min},1]}
\left[
\sqrt{
\frac{
\lambda_j\,\widetilde s_j(W)
}{
\eta\,\widetilde d_j^2\,c_j(W)
}
}
\right].
\label{eq:allocation}
\end{equation}
The clipping, $\operatorname{clip}_{[a,b]}(x)=\min\{\max(x,a),b\}$, keeps every replay probability in $[q_{j,\min},1]$. Theorem~\ref{thm:allocation} shows that this balance point is the exact solution of \eqref{eq:minimax}, not an approximation to it.

Equation~\eqref{eq:allocation} shows where the budget goes. Within a baseline, the quantities that vary from task to task decide which tasks are replayed: replay rises where $\widetilde s_j(W)$ is large, because that outcome is harder to predict, and falls where $c_j(W)$ is large, because replay is more expensive. Across the two baselines, $\lambda_j$ and $\widetilde d_j^2$ shift the budget toward the comparison that needs more evidence. Only $\lambda_H$ and $\eta$ are left to set. For a given $\lambda_H$, with $\lambda_A=1-\lambda_H$, the multiplier $\eta$ is adjusted until the two rules together spend exactly the budget $B$ of \eqref{eq:budget}. \TEAM\ scans $\lambda_H$ over $[0,1]$, computes both estimated difficulties at each value, and keeps the pair of rules whose larger difficulty is smallest. Both are one-dimensional searches, so no solver is needed.

\paragraph{Making the final decision.}
For each evaluation task $i$, write $q_{ij}^\star=q_j^\star(W_i)$, draw $R_{iH}$ and $R_{iA}$ separately from $\mathrm{Bernoulli}(q_{iH}^\star)$ and $\mathrm{Bernoulli}(q_{iA}^\star)$, and record both probabilities. Equation~\eqref{eq:margin-estimator} then gives $\widehat\Delta_H$ and $\widehat\Delta_A$. Letting $\widehat\sigma_j^2$ be the empirical variance of the scores $\psi_{1j},\ldots,\psi_{nj}$, the one-sided lower confidence bound is $L_j=\widehat\Delta_j-z_{1-\alpha}\widehat\sigma_j/\sqrt n$, where $z_{1-\alpha}$ is the $(1-\alpha)$ quantile of the standard normal distribution and $\alpha=0.05$.

The human--AI workflow is declared to beat both alternatives only when
\begin{equation}
L_H>\delta_H
\quad\text{and}\quad
L_A>\delta_A.
\label{eq:decision}
\end{equation}
Requiring both bounds to clear their thresholds needs no multiplicity correction, for the reason given with Theorem~\ref{thm:inference}. Algorithm~\ref{alg:team} states the procedure end to end, and Section~\ref{sec:theory} establishes that it remains valid when the rules are chosen from the pilot.

\begin{algorithm}[!ht]
\caption{\TEAM\ pseudocode}
\label{alg:team}
\footnotesize
\begin{algorithmic}[1]
\Require Independent pilot of $n_{\mathrm{p}}$ fully observed tasks; evaluation pool $\{(W_i,T_i)\}_{i=1}^n$; replay costs $c_j(W_i)$; budget $B$; floors $q_{j,\min}$; stabilizer $\tau$; thresholds $\delta_H,\delta_A$.
\State From the pilot, estimate $\widetilde\Delta_j$, $\widetilde m_j(W)$, $\widetilde s_j(W)$, and $\widetilde\kappa_j$ (Appendix~\ref{app:variance}), for $j\in\{H,A\}$.
\State Compute the stabilized gap $\widetilde d_j^2$ from \eqref{eq:stabilized-gap}.
\State Search over $\lambda_H\in[0,1]$, with $\lambda_A=1-\lambda_H$, and for each bisect $\eta$ to meet the budget $B$, giving $q_{ij}^\star$ from \eqref{eq:allocation}. Appendix~\ref{app:splits} gives the budget range over which such an $\eta$ exists.
\State Keep the $\lambda_H$ that minimizes the larger of the two estimated difficulties.
\State For each task $i$, draw $R_{iH}\sim\mathrm{Bernoulli}(q_{iH}^\star)$ and $R_{iA}\sim\mathrm{Bernoulli}(q_{iA}^\star)$ separately, replay baseline $j$ only when $R_{ij}=1$, and record every $q_{ij}^\star$.
\State Compute $\widehat\Delta_H,\widehat\Delta_A$ from \eqref{eq:margin-estimator} and report the decision \eqref{eq:decision}.
\end{algorithmic}
\end{algorithm}

\section{Theoretical guarantees}
\label{sec:theory}

Section~\ref{sec:team-design} gave a practical rule for assigning replay probabilities. Three results justify it. First, we show why the harder comparison determines the approximate number of evaluation tasks needed. Second, we show that the square-root rule of \eqref{eq:allocation} solves the minimax problem of Section~\ref{sec:problem}. Third, we show that choosing the replay probabilities from the pilot and then drawing replays at random still controls the chance of wrongly declaring that the workflow beats both baselines.

\begin{proposition}[The harder comparison determines evaluation effort]
\label{prop:effort}
Assume both gains exceed their required thresholds and the two gain estimates are asymptotically normal. To obtain one-sided level-$\alpha$ tests with joint power at least $1-\beta$, it is sufficient, to first order, that
\begin{equation}
n
\ge
\{z_{1-\alpha}+z_{1-\beta/2}\}^2
\max_{j\in\{H,A\}}
\frac{\sigma_j^2(q_j)}
     {(\Delta_j-\delta_j)^2}.
\label{eq:effort}
\end{equation}
Therefore, minimizing the objective in Section~\ref{sec:problem} minimizes this first-order sample-size requirement.
\end{proposition}

The result explains the maximum in our design objective: the comparison requiring more evidence determines the evaluation effort. Appendix~\ref{app:effortproof} proves it by asking each comparison to succeed with probability $1-\beta/2$, so that both hold with probability at least $1-\beta$. The bound says how many tasks are enough, not exactly how often both bounds will clear at a given budget, and Section~\ref{sec:scope} measures the difference.

\begin{theorem}[\TEAM\ solves the replay problem]
\label{thm:allocation}
Suppose $\Delta_j>\delta_j$ for $j\in\{H,A\}$, the replay costs $c_j(W)$ are nonnegative and finite, and the budget is feasible. On tasks with $c_j(W)=0$ take $q_j^\star(W)=1$. Where $c_j(W)>0$, an optimal replay rule for the cost-constrained design problem of Section~\ref{sec:problem} has the form
\begin{equation}
q_j^\star(W)
=
\operatorname{clip}_{[q_{j,\min},1]}
\left[
\sqrt{
\frac{\lambda_j s_j(W)}
     {\eta(\Delta_j-\delta_j)^2c_j(W)}
}
\right],
\label{eq:allocation-true}
\end{equation}
for multipliers $\lambda_H,\lambda_A\ge0$ with $\lambda_H+\lambda_A=1$ and a budget multiplier $\eta>0$ when the budget binds. Complementary slackness gives positive weight only to a comparison that attains the maximum in the design objective, so both multipliers are positive where the budget lets the two comparisons equalize.

Replacing the unknown quantities by the pilot estimates gives the \TEAM\ rule in \eqref{eq:allocation}.
\end{theorem}

The square-root rule is therefore not a heuristic but the exact solution; Appendix~\ref{app:allocation} proves it from the KKT conditions of the convex program.

\begin{theorem}[Valid inference after pilot-based replay]
\label{thm:inference}
Write $\mathcal P$ for the pilot. Suppose $\mathcal P$ is independent of the final evaluation pool, the replay probabilities are fixed before the missing baseline outcomes are observed, and $q_j(W)\ge q_{j,\min}>0$. Assume the tasks are independent and identically distributed, with outcomes bounded and $\|\widetilde m_j\|_\infty<\infty$ for almost every $\mathcal P$. Assume also that, given $\mathcal P$, the population multiplier search has a unique solution $q^\circ$ (Appendix~\ref{app:inferenceproof}). Then, conditionally on $\mathcal P$,
\begin{equation}
\sqrt n
\begin{pmatrix}
\widehat\Delta_H-\Delta_H\\
\widehat\Delta_A-\Delta_A
\end{pmatrix}
\;\Big|\;\mathcal P
\xrightarrow{d}
N\{0,\Sigma_{\mathcal P}\},
\label{eq:joint-normal}
\end{equation}
with $\Sigma_{\mathcal P}$ the covariance at the limiting design $q^\circ$ that the pilot induces. Consequently the decision rule \eqref{eq:decision} has conditional asymptotic false-claim probability at most $\alpha$ whenever at least one true gain fails its threshold; averaging over $\mathcal P$ gives the same bound unconditionally.
\end{theorem}

The reason is that the pilot fixes the replay probabilities before any evaluation outcome is observed, and the estimator records those probabilities when correcting for selective replay. The pilot therefore sets efficiency, not validity: a poor pilot gives a poor design and hence wider intervals, but leaves the error rate untouched. A false claim needs both bounds to clear while one comparison truly fails, so no multiplicity adjustment is needed; formally, this is an intersection--union test \cite{berger1982,laska1989}.

\section{Results}
\label{sec:results}

We first ask whether the two-baseline deployment question matters in real data. We then test \TEAM\ itself: Section~\ref{sec:alloc} uses simulation, where the true gains are known and we control which baseline outcomes are revealed, and Section~\ref{sec:semisynth} keeps the dependence and variability of real chest X-ray data. Section~\ref{sec:scope} asks when the method helps and when it does not. Within each replication every method sees the same pilot, evaluation pool, and potential outcomes, following the protocol set out in Appendix~\ref{app:simulationplan}. Unless stated otherwise, we use $\delta_H=\delta_A=0$.

\subsection{Do clinical human--AI workflows beat both alternatives?}
\label{sec:clinical}

\paragraph{Intuition.}
The two-comparison decision matters only if comparing the human--AI workflow with the human alone can give a different answer from comparing it with the agent alone. We first ask whether this happens in real clinical data.

\paragraph{Setting.}
We reanalyze two studies in which clinicians evaluate the same cases with and without AI support \cite{collabcxr,chanda2024}. The model prediction is also recorded, so each case provides outcomes for the assisted clinician, the clinician alone, and the model alone. Each outcome is a Brier reward \cite{brier1950}: if a clinician or model states probability $p$ for a case whose true label is $y\in\{0,1\}$, the reward is $1-(p-y)^2$, so higher is better.

\paragraph{Finding.}
Across \RClinCells\ clinical settings and \RClinReads\ reads, each a single clinician judging a single case, the assisted clinician performs better than the clinician alone in every setting, but worse than the model alone in every setting. In no setting does the assisted clinician beat both alternatives (\RClinBeatsBoth\ of \RClinCells). Table~\ref{tab:clinical} gives the two gains per setting with reader-clustered standard errors, and marks which lower bounds clear zero. Because the same clinicians evaluate some cases more than once, seeing a case in one condition may affect a later judgment of it.

\begin{table}[!ht]
\centering
\small
\setlength{\tabcolsep}{4.5pt}
\caption{Matched human--AI reader studies judged against \emph{both} baselines, with reader-clustered standard errors in parentheses. \textbf{Bold} marks a one-sided 95\% lower bound above zero. In no setting does the assisted clinician beat both alternatives.}
\label{tab:clinical}
\begin{tabular}{lrcc c}
\toprule
Study & Reads & $\Delta_H$ (vs.\ clinician) & $\Delta_A$ (vs.\ AI) & Beats both \\
\midrule
Chest X-ray, design 2, no history & 1,980 & \textbf{+0.0056 (0.0026)} & -0.0175 (0.0055) & no \\
Chest X-ray, design 2, with history & 1,977 & +0.0024 (0.0031) & -0.0143 (0.0063) & no \\
Chest X-ray, design 3, no history & 875 & \textbf{+0.0185 (0.0073)} & -0.0246 (0.0091) & no \\
Chest X-ray, design 3, with history & 875 & +0.0106 (0.0076) & -0.0157 (0.0073) & no \\
Melanoma, AI support & 1,508 & \textbf{+0.0836 (0.0107)} & -0.0206 (0.0108) & no \\
Melanoma, AI with explanations & 1,508 & \textbf{+0.0713 (0.0107)} & -0.0328 (0.0108) & no \\
\bottomrule
\end{tabular}
\end{table}

\paragraph{Contrast.}
A coding benchmark orders the three configurations the other way, with pass rates of 0.67\% for the agent alone and 18.89\% for the human alone against 31.11\% for the workflow \cite{haieval2025}, though only in aggregate, without the task-level pairing our two comparisons need. Improvement over the human alone does not settle the deployment question.

\subsection{Does decision-targeted replay help when one comparison is harder?}
\label{sec:alloc}

\paragraph{Intuition.}
\TEAM\ is intended to help when one of the two comparisons requires substantially more evidence than the other. Replay does not change the true gains, only how precisely we see them. We test whether directing replay toward the harder comparison lets both gains clear their thresholds more often at the same cost.

\paragraph{Setting.}
Each replication draws a fully observed pilot and a separate evaluation pool, then hides the pool's human-only and agent-only outcomes. \TEAM\ sets the two replay probabilities, and only the drawn replays are revealed. Tasks fall into two trace groups of differing prediction difficulty, and five scenarios vary how close the two gains sit to their thresholds and how much each replay costs (Appendix~\ref{app:dgp}). Two comparators run here: Neyman allocation, which uses prediction uncertainty and cost but not distance from the threshold, and uniform replay, which becomes an equal cost split in the unequal-cost scenario. Table~\ref{tab:results} adds a surrogate oracle, the same rule given the true gains in place of pilot estimates. The chest X-ray study below adds two more, \emph{AI-uncertainty} replay, which buys where the model is least certain, and \emph{constant $q$}, which replays every case at one rate. Appendix~\ref{app:baselines} defines each. We report how often both lower confidence bounds clear zero.

\begin{figure}[!ht]
\centering
\includegraphics[width=0.52\linewidth]{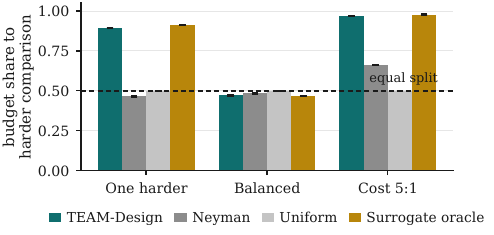}
\caption{How the replay budget divides between the two baselines, aggregated from the task-level probabilities $q_{iH}$ and $q_{iA}$. \TEAM\ raises the replay probabilities for the harder comparison when one clearly requires more evidence, and stays closer to an equal split otherwise.}
\label{fig:alloc}
\end{figure}

\paragraph{Finding.}
When one comparison sits much closer to its threshold, \TEAM\ concludes that the workflow beats both baselines in \RBotTeam\% of replications against \RBotNeyman\% for Neyman at the same cost. It gets there by raising the replay probabilities for the harder comparison, and settles near an equal split when neither is harder (Figure~\ref{fig:alloc}). The advantage survives replay costs that differ by five to one, and reverses when the two comparisons are equally hard, where Neyman does better because there is no imbalance to exploit. At the decision boundary \TEAM's false-claim rate stays below 5\%, even when the pilot is deliberately misled. Table~\ref{tab:results} gives all five scenarios with Monte Carlo standard errors.

\begin{table}[!ht]
\centering
\small
\setlength{\tabcolsep}{4.5pt}
\caption{Beats-both decision rate: replications (\%) with both lower bounds above zero; under the boundary null this is a false-claim rate at a nominal 5\%. Monte Carlo errors in parentheses, $R=20{,}000$ per row. The uniform column is an equal cost split in the unequal-cost row. The oracle allocates from the true gains, a reference design and not a deployable rule.}
\label{tab:results}
\begin{tabular}{lcccc}
\toprule
Setting & \TEAM & Neyman & Uniform & Surrogate oracle \\
\midrule
One harder comparison & 77.1 (0.30) & 65.1 (0.34) & 65.8 (0.34) & 79.8 (0.28) \\
Balanced comparisons & 39.1 (0.35) & 44.4 (0.35) & 43.7 (0.35) & 44.3 (0.35) \\
Unequal replay cost (5:1) & 57.9 (0.35) & 54.1 (0.35) & 45.2 (0.35) & 56.2 (0.35) \\
\midrule
Boundary null & 3.6 (0.13) & 4.6 (0.15) & 4.8 (0.15) & 3.4 (0.13) \\
Boundary null, reversed pilot & 3.7 (0.13) & 4.5 (0.15) & 4.7 (0.15) & 3.3 (0.13) \\
\bottomrule
\end{tabular}
\end{table}

\subsection{Validation on real clinical structure}
\label{sec:semisynth}

\paragraph{Intuition.}
In the real studies the model alone beats the assisted clinician by margins far larger than their standard errors, so the answer is the same however the budget is spent. A design only matters when the gains sit close to their thresholds. We therefore keep the real cases, costs, and readers, and shift only the average outcomes so the gains land there.

\paragraph{Setting.}
The shift makes the agent comparison the harder one. Here the model's output costs nothing and only clinician reads are purchased, so this study does not exercise the cost trade-off between the two baselines; Section~\ref{sec:alloc} remains the direct test of that. A quarter of the cases are held out as the pilot and the rest form the evaluation pool. Because the split is by case, the same readers appear in both, so Theorem~\ref{thm:inference} does not apply exactly here. Appendix~\ref{app:semisynth} states which quantities keep their measured values and which are shifted.

\begin{figure}[!ht]
\centering
\includegraphics[width=0.48\linewidth]{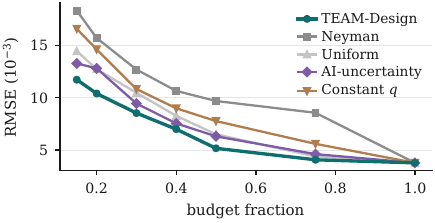}
\caption{Root mean squared error on the harder comparison as a function of the replay budget, on the structure of the chest X-ray study. All methods coincide at full replay.}
\label{fig:semisynth}
\end{figure}

\paragraph{Finding.}
\TEAM\ has the lowest root mean squared error for estimating the gain against the harder baseline at every budget below full replay (Figure~\ref{fig:semisynth}); at one fifth of full replay it reaches $\RSSLoTeam\times10^{-3}$ against $\RSSLoNeyman\times10^{-3}$ for Neyman, where that gain is itself about $\RSSTruthA\times10^{-3}$. Lower error does not always mean both comparisons pass more often: the true agent gain sits so close to its threshold that a noisier estimate crosses it more often by chance, which is why we compare on error here. Appendix~\ref{app:floor} reports a floor sweep that leaves the pattern unchanged.

\subsection{Scope conditions}
\label{sec:scope}

\begin{figure}[!ht]
\centering
\includegraphics[width=0.74\linewidth]{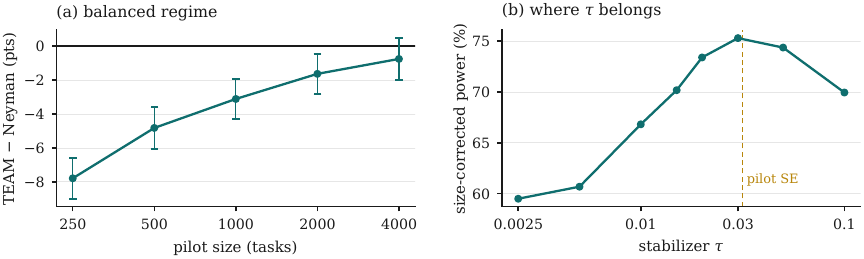}
\caption{Scope of \TEAM. \textbf{(a)} When the two comparisons are similarly difficult, \TEAM\ trails Neyman, with the gap shrinking as the pilot grows. \textbf{(b)} Performance against the stabilizer $\tau$, with the pilot standard error marked.}
\label{fig:scope}
\end{figure}

Two things bound where \TEAM\ helps. The first is imbalance: it needs a clearly harder comparison and enough budget to shift measurements to it, so when the two are similarly difficult nothing is left to exploit and Neyman does better, though a larger pilot narrows the gap (Figure~\ref{fig:scope}a).

The second is the objective, which measures large-sample difficulty and not the exact chance that both bounds clear at a fixed budget; a noisy pilot that puts one comparison very close to its threshold can therefore draw too much replay. The stabilizer $\tau$ holds that in check and works best on the scale of the pilot's uncertainty (Figure~\ref{fig:scope}b). \TEAM\ is therefore a decision-targeted design for unequal comparisons, not a uniformly better replay rule.

Appendix~\ref{app:gaussian} measures the gap against Bayesian designs aimed at the decision, and Appendix~\ref{app:sens} reports how the results respond to the stabilizer, the replay floor, the budget, and a non-zero $\delta_j$.

\section{Conclusion}
\label{sec:conclusion}

Keeping a human--AI workflow in production means beating both alternatives at once, and neither is observed once the workflow runs. Measuring either costs expert time, so we made what to buy part of the design. \TEAM\ assigns each task a human-only and an agent-only replay probability, giving more measurement to the comparison that needs more evidence while accounting for prediction difficulty and replay cost. In the clinical data the assisted clinician beats the clinician alone but not the model alone, so evaluating only the human comparison can give the wrong deployment conclusion.

\paragraph{Limitations.}
The rule helps where one comparison is clearly harder, not where the two are balanced, and its objective bounds joint power to first order instead of maximizing it at a fixed budget. The clinical evidence is matched reader studies, not deployments.

\paragraph{Reproducibility.}
\url{https://github.com/HamedKhosravi99/team-design-code}

\clearpage
\appendix
\section{Proofs}
\label{app:proofs}

\subsection{Proof of Proposition~\ref{prop:effort}}
\label{app:effortproof}
For comparison $j$, asymptotic normality gives
\begin{equation}
\frac{\sqrt n(\widehat\Delta_j-\Delta_j)}{\sigma_j(q_j)}\xrightarrow{\ d\ } N(0,1).
\end{equation}
A one-sided level-$\alpha$ Wald test rejects $H_{0j}:\Delta_j\le\delta_j$ when
\begin{equation}
\widehat\Delta_j-\delta_j>z_{1-\alpha}\frac{\sigma_j(q_j)}{\sqrt n}.
\end{equation}
Under a fixed positive alternative its asymptotic power is
\begin{equation}
\Phi\!\left(\sqrt n\frac{\Delta_j-\delta_j}{\sigma_j(q_j)}-z_{1-\alpha}\right).
\end{equation}
Requiring this to be at least $1-\beta/2$ gives
\begin{equation}
\sqrt n\frac{\Delta_j-\delta_j}{\sigma_j(q_j)}\ge z_{1-\alpha}+z_{1-\beta/2}.
\end{equation}
Squaring and requiring the inequality for both comparisons yields \eqref{eq:effort}. \hfill$\square$

\subsection{Variance decomposition for Equation~(\ref{eq:variance})}
\label{app:variance}
Write the score of \eqref{eq:score} for a generic task, dropping the index $i$, as
\begin{equation}
\psi_j=(T-Y_j)+\left(1-\frac{R_j}{q_j(W)}\right)\{Y_j-\widetilde m_j(W)\}.
\end{equation}
Conditional on the pilot $\mathcal P$, replay is randomized using the recorded task information $W$ alone, so $R_j\perp(T,Y_j)\mid(W,\mathcal P)$ and $\E[R_j\mid W,T,Y_j,\mathcal P]=q_j(W)$. The second term therefore has mean zero conditional on $(W,T,Y_j)$, so it is uncorrelated with the first. Its conditional variance is $\{Y_j-\widetilde m_j(W)\}^2\Var\{R_j/q_j(W)\mid W\}=\{Y_j-\widetilde m_j(W)\}^2\{1-q_j(W)\}/q_j(W)$. Taking expectations,
\begin{equation}
\sigma_j^2(q_j)=\Var(T-Y_j)+\E\left[\frac{s_j(W)}{q_j(W)}\right]-\E[s_j(W)],
\end{equation}
which is \eqref{eq:variance} with $\kappa_j=\Var(T-Y_j)-\E[s_j(W)]$. No assumption relating $Y_j$ to $T$ given $W$ is used, so the decomposition holds under the correlated outcomes of Appendix~\ref{app:dgp}. The pilot estimate $\widetilde\kappa_j$ of Section~\ref{sec:team-design} targets this quantity by replacing both population moments with their pilot counterparts, $\widetilde\kappa_j=\Varp(T-Y_j)-\Ep[\widetilde s_j(W)]$, where $\Varp$ is the pilot sample variance, which the pilot can form because all three outcomes are observed there. The simpler $\Var\{T-\widetilde m_j(W)\}$ agrees with it only for a well-specified predictor.

\hfill$\square$

\subsection{Proof of Theorem~\ref{thm:allocation}}
\label{app:allocation}
For $j\in\{H,A\}$ write
\[
u_j(W)=\frac{s_j(W)}{(\Delta_j-\delta_j)^2},
\qquad
b_j=\frac{\kappa_j}{(\Delta_j-\delta_j)^2},
\]
so that by \eqref{eq:variance} comparison $j$'s difficulty is $b_j+\E[u_j(W)/q_j(W)]$ and the epigraph constraint that it be at most $t$ reads
\begin{equation}
b_j+\E\left[\frac{u_j(W)}{q_j(W)}\right]\le t.
\end{equation} Introduce multipliers $\lambda_H,\lambda_A\ge0$ for the comparison constraints and $\eta\ge0$ for the budget, positive when it binds. The Lagrangian terms involving $t$ and $q$ are
\begin{equation}
\mathcal L
=t+\sum_j\lambda_j\left\{b_j+\E\left(\frac{u_j}{q_j}\right)-t\right\}
+\eta\{\E(c_Hq_H+c_Aq_A)-B\}.
\end{equation}
Stationarity in $t$ gives $\lambda_H+\lambda_A=1$. At an interior point,
\begin{equation}
-\frac{\lambda_j u_j(W)}{q_j(W)^2}+\eta c_j(W)=0,
\end{equation}
so
\begin{equation}
q_j(W)=\sqrt{\frac{\lambda_j u_j(W)}{\eta c_j(W)}},
\end{equation}
which is \eqref{eq:allocation-true}. The box constraints clip this expression to $[q_{j,\min},1]$. Convexity of $u/q$ on $q>0$ makes the KKT conditions sufficient, and Slater's condition, which the feasible budget range of Appendix~\ref{app:splits} supplies, makes them necessary, so an optimal rule takes this form. Where $c_j(W)=0$ the per-task term is $\lambda_ju_j(W)/q_j(W)$ alone, which is strictly decreasing, so the box constraint carries $q_j(W)$ to one and no stationary point is required. Complementary slackness yields the limiting-comparison statement. Replacing $s_j(W)$ and $(\Delta_j-\delta_j)^2$ by their pilot counterparts $\widetilde s_j(W)$ and $\widetilde d_j^2$ gives the implementable rule \eqref{eq:allocation}. \hfill$\square$

\subsection{Proof of Theorem~\ref{thm:inference}}
\label{app:inferenceproof}
Write $\mathcal F_n$ for the $\sigma$-field generated by the pilot, the pool covariates $W_{1:n}$ and the costs. By hypothesis $q_H$ and $q_A$ are $\mathcal F_n$-measurable and no baseline outcome or replay indicator enters them.

\paragraph{The conditional mean.}
Given $\mathcal F_n$ the indicators are independent across tasks and comparisons, with $R_{ij}\sim\mathrm{Bernoulli}\{q_j(W_i)\}$ independent of $(T_i,Y_{iH},Y_{iA})$ given $W_i$, so the correction term of \eqref{eq:score} has conditional mean zero and
\begin{equation}
\E[\psi_{ij}\mid\mathcal F_n]=\mu_j(W_i),\qquad \mu_j(W)=\E[T-Y_j\mid W].
\label{eq:condmean}
\end{equation}
The conditioning centers each score at the \emph{conditional} gain, not at $\Delta_j$, and the gap between them is carried by the second term below.

\paragraph{A decomposition that isolates the design.}
\begin{equation}
\sqrt n\,(\widehat\Delta_j-\Delta_j)
=\underbrace{\frac{1}{\sqrt n}\sum_i\{\psi_{ij}-\mu_j(W_i)\}}_{A_{nj}}
+\underbrace{\frac{1}{\sqrt n}\sum_i\{\mu_j(W_i)-\Delta_j\}}_{B_{nj}} .
\label{eq:decomp}
\end{equation}
Every dependence on $q_j$ sits in $A_{nj}$. The term $B_{nj}$ is a normalized sum of independent mean-zero variables that the design cannot touch.

\paragraph{The design-dependent term.}
Conditionally on $\mathcal F_n$ the summands of $A_{nj}$ are independent with mean zero by \eqref{eq:condmean}. Outcomes are bounded, the realized pilot gives $\|\widetilde m_j\|_\infty<\infty$, and $q_j\ge q_{j,\min}>0$, so the summands are bounded by a fixed multiple of $1/q_{j,\min}$ and the Lindeberg condition for the triangular array holds. Their average conditional variance is
\begin{equation}
\frac1n\sum_i\left[\Var(T-Y_j\mid W_i)+s_j(W_i)\Big\{\frac{1}{q_j(W_i)}-1\Big\}\right].
\end{equation}

\paragraph{The design converges given the pilot.}
Fix $\mathcal P$. Conditional on the pilot, $\widetilde s_j$, $\widetilde d_j^2$ and $\widetilde\kappa_j$ are fixed, and the remaining ingredients of the estimated difficulty of comparison $j$ are pool averages. Because the replay probabilities are bounded below by $q_{j,\min}>0$ and the multiplier search runs over a compact set, the empirical comparison objectives converge uniformly in probability to their conditional population counterparts. Under the assumed uniqueness of the population minimizer, the argmin theorem gives consistency of the selected multipliers and, writing $q_{nj}$ for the rule the pool of size $n$ selects, $q_{nj}\to q_j^{\circ}$ in probability, uniformly in $W$ by continuity of \eqref{eq:allocation}, where $q_j^{\circ}$ is $\mathcal P$-measurable. The displayed average therefore converges in probability to $\E[\Var(T-Y_j\mid W)]+\E[s_j/q_j^{\circ}]-\E[s_j]$.

\paragraph{The limit.}
Work with the pairs $A_n=(A_{nH},A_{nA})$ and $B_n=(B_{nH},B_{nA})$, so that the covariance between the two comparisons is carried along. The conditional multivariate Lindeberg--Feller theorem gives $\E[\exp(it^\top A_n)\mid\mathcal F_n]\to\exp(-\tfrac12 t^\top\Sigma_A t)$ in probability. Since $B_n$ is $\mathcal F_n$-measurable,
\[
\E\!\left[e^{it^\top A_n+iu^\top B_n}\right]
=\E\!\left[e^{iu^\top B_n}\,\E\{e^{it^\top A_n}\mid\mathcal F_n\}\right],
\]
whose inner factor converges to a nonrandom limit, while the ordinary multivariate central limit theorem gives $B_n\Rightarrow N(0,\Sigma_B)$. The joint characteristic function therefore converges to the product of two Gaussian characteristic functions, so $A_n$ and $B_n$ are asymptotically independent given the pilot and $A_n+B_n\Rightarrow N(0,\Sigma_A+\Sigma_B)$, with variance
\begin{equation}
\E[\Var(T-Y_j\mid W)]+\Var\{\mu_j(W)\}+\E\!\left[\frac{s_j}{q_j^{\circ}}\right]-\E[s_j]
=\kappa_j+\E\!\left[\frac{s_j}{q_j^{\circ}}\right],
\end{equation}
by the variance decomposition of $T-Y_j$ over $W$. That is \eqref{eq:variance} at the limiting design, so the limit is $N\{0,\Sigma(q_H^{\circ},q_A^{\circ})\}$. The empirical covariance of the paired scores is consistent by the same boundedness.

\paragraph{The reported rate.}
If $\Delta_j\le\delta_j$ for some $j$, then $\{L_H>\delta_H\}\cap\{L_A>\delta_A\}\subseteq\{L_j>\delta_j\}$, and the right-hand event is the rejection of a true one-sided component null, whose asymptotic probability is at most $\alpha$. \hfill$\square$

\section{Simulation and empirical study design}
\label{app:simulationplan}

Section~\ref{sec:results} reports three constructed studies and one reanalysis of published reader data. This appendix gives the construction of the three constructed studies, the information available to every competing rule, and the protocol under which the sweeps were run. The reanalysis needs no construction, since it uses the published reader studies as they stand, and Section~\ref{sec:clinical} states how its outcomes are scored.

\subsection{Generating the evaluation data}
\label{app:dgp}

Measuring how much error a replay rule leaves requires knowing the true gains, which no real study reports. We therefore generate tasks whose gains are fixed in advance. The scenarios below differ only in how unequal the two comparisons are, since that imbalance is what the design is built to exploit.

\paragraph{Unit and estimand.}
One unit is one software task. Each task has a trace stratum $W\in\{0,1\}$, a deployed human--AI workflow outcome $T$, a human-only baseline outcome $Y_H$, and an agent-only baseline outcome $Y_A$. The primary outcome is binary task success. The estimands are exactly the two population gains \eqref{eq:margins}, and no scenario changes the estimand.

\paragraph{Primary two-stratum generator.}
Here $W\sim\mathrm{Bernoulli}(1/2)$, and conditional on $W$ the three outcomes $T,Y_H,Y_A$ are drawn independently as Bernoulli variables with stratum-specific success probabilities. The primary one-bottleneck specification, the harder-comparison scenario of Table~\ref{tab:results}, is
\begin{equation}
P(T=1\mid W)=(0.70,0.78),\;
P(Y_H=1\mid W)=(0.55,0.85),\;
P(Y_A=1\mid W)=(0.60,0.56).
\label{eq:dgp-primary}
\end{equation}
This gives a small workflow-versus-human gain and a substantially larger workflow-versus-agent gain while making human-only residual uncertainty strongly stratum dependent. These values are design parameters, not empirical findings.

\paragraph{Balanced-comparison specification.}
This specification keeps $T$ and $Y_H$ as in Equation~\eqref{eq:dgp-primary} and replaces the agent baseline by
\begin{equation}
P(Y_A=1\mid W)=(0.72,0.68),
\end{equation}
so that the two population gains are of comparable size. This is the pre-specified balanced-comparison check, since a method built around a unique limiting comparison should have little structural advantage here.

\paragraph{Boundary-null specification.}
A one-sided test claims a win most readily when the true gain sits exactly at its threshold, and a false claim needs only one of the two comparisons to fail. The least favorable case is therefore a null on one comparison with the other left comfortably positive, which is what this specification builds. It sets
\begin{equation}
P(T=1\mid W)=(0.66,0.74)
\end{equation}
with the original $Y_H$ and $Y_A$ probabilities, making the workflow-versus-human gain exactly zero while the workflow-versus-agent gain remains positive. This is the primary false-claim experiment.

\paragraph{Correlation sensitivity.}
The independent-Bernoulli generator is transparent but does not create residual correlation between potential outcomes after conditioning on $W$. Every primary scenario is therefore repeated with a Gaussian-copula generator, which draws $(Z_T,Z_H,Z_A)$ from a trivariate normal with pairwise correlation $\rho\in\{0.25,0.50,0.75\}$ and thresholds each coordinate to reproduce the same Bernoulli marginals. This tests whether the allocation result depends on conditional independence.

\paragraph{Unequal-cost scenario.}

A human-only replay and an agent-only replay rarely cost the same. Recovering the human baseline can take a senior engineer or a clinician the better part of an hour, while recovering the agent baseline costs compute, so a rule blind to that difference will spend its budget on whichever evidence is cheap and not on whichever evidence the decision needs. This scenario prices the two apart. It keeps the primary bottleneck generator and sets the cost ratio
\begin{equation}
c_H/c_A=5,
\end{equation}
with the expected total cost fixed at one cost unit per evaluation task. \TEAM, cost-aware Neyman, and equal cost splitting are compared at matched \emph{expected} cost, and the primary result is the beats-both decision rate. The secondary results are the number of human-only replays, the number of agent-only replays, the gain standard errors, and the cost per successful beats-both decision.

The ratio is also swept over $c_H/c_A\in\{2,5,10\}$. The main text reports only the pre-specified $5{:}1$ setting, and the full curve is Table~\ref{tab:sensitivity}.

\paragraph{Real clinical structure.}
\label{app:semisynth}

A generator supplies known gains but invents the correlation between readers and cases. The chest X-ray study supplies that correlation but places its gains far enough above their thresholds that no replay rule can change the answer. This experiment keeps the study's covariates, costs, reader structure and correlations, and shifts only the arm means.
Every measured feature of the chest X-ray study is retained, namely the covariates, the per-case radiologist times that supply the replay costs, the reader and case structure used for clustering, the residual heteroskedasticity, and the within-case correlation. Only the arm means are shifted, which changes no second moment. The reported regime makes the agent comparison the limiting one at $(\Delta_H,\Delta_A)=(0.08,0.01)$, and the balanced regime $(0.04,0.04)$ and a boundary regime $(0.05,0.00)$ are run alongside it.

A quarter of cases are held out as the fully observed pilot and the rest form the evaluation pool, about $656$ reads. Splitting by case and not by reader leaves readers common to both, so the independence of Theorem~\ref{thm:inference} holds across cases and not across readers, which is why every standard error here is reader-clustered. Its i.i.d. hypothesis is therefore met at the case level and not at the read level, so this experiment sits outside the theorem's exact assumptions and its false-claim rates are an empirical check and not an instance of the guarantee. Replay cost is the measured per-case radiologist time, so a budget of $1.0$ is the cost of replaying every case. The sweep runs $0.15$ to $1.0$ and the pilot is not charged against it, being already collected. The probability floor is $q_{j,\min}=0.05$ and each point is $\RSSReps$ replications. The stabilizer is set separately for each comparison to the pilot's own standard error of that gain, which is the scale Section~\ref{sec:scope} finds works best.

Five rules are compared. \TEAM\ is the square-root rule of \eqref{eq:allocation}, which prices prediction difficulty against replay cost and tilts the budget toward the comparison that is harder to settle. \emph{Neyman} weights by residual variance and cost but ignores the gains. \emph{Uniform} uses one constant probability for both baselines at the same expected cost. \emph{Constant $q$} uses one constant probability at the same expected \emph{count} of replays. \emph{AI-uncertainty} weights by the model's own predictive uncertainty on the case, the heuristic a practitioner would reach for first. The reported quantity is the root mean squared error of the limiting gain.

The agent-limited and balanced regimes are chosen to differ only in how unequal the comparisons are. The agent-limited regime sets the agent gain at $0.01$ against a human gain of $0.08$, so the agent comparison needs far more evidence, and the balanced regime sets both at $0.04$. Replay costs are the study's own measured times, an assisted read at $117$\,s against an unaided read at $176$\,s.

The agent-limited regime sits at the edge of what the data can resolve. At full replay, where every case is bought and no design choice remains, the one-sided bound on the agent gain needs $\RSSThresh\times10^{-3}$ while the true gain is $\RSSTruthA\times10^{-3}$, so both bounds clear in only $\RSSFullDec$\% of replications and every rule scores alike.

\paragraph{Protocol.}
\label{app:splits}

Theorem~\ref{thm:inference} requires the pilot to share no task with the evaluation pool and every replay probability to be fixed before any baseline outcome is observed. Neither condition leaves a trace in the results when it is broken. Table~\ref{tab:settings} gives the sizes and the budgets, and the paragraphs below give the protocol under which each replication was run.

\begin{table}[!ht]
\centering
\small
\caption{Settings for the three constructed studies. The synthetic and semi-synthetic sizes count tasks and reads, and the Gaussian sizes count measurements.}
\label{tab:settings}
\begin{tabular}{lccc}
\toprule
 & Synthetic & Semi-synthetic & Gaussian \\
\midrule
Pilot size & $500$ & a quarter of cases & a tenth of the budget \\
Evaluation size & $3{,}000$ & about $656$ & $18{,}000$ \\
Replay budget $B$ & $0.4$ & $0.15$ to $1.0$ & --- \\
Probability floor $q_{j,\min}$ & $0.02$ & $0.05$ & --- \\
Stabilizer $\tau$ & $0.03$ & pilot standard error & $0.03$ \\
Replications & $20{,}000$ & $\RSSReps$ & $30{,}000$ \\
\bottomrule
\end{tabular}
\end{table}

\paragraph{Monte Carlo protocol.}

The bottleneck, balanced, boundary-null, and cost-ratio scenarios use 20{,}000 replications, and the correlation and size sweeps use 2{,}000. Unit tests confirm that every method meets the same expected or realized budget, that on fully observed data the estimator recovers the known simulated gains, and that setting all $q_j=1$ reproduces the complete-data estimator. Each replication saves one row holding the true and estimated gains, their standard errors, replay counts and cost, allocation shares, and the final beats-both decision, and every reported table derives from those rows.

All experiments run on a single CPU machine with no accelerator. The full replay grid takes about twelve minutes and the matched-validity comparison about six, with the clinical reanalyses and the semi-synthetic sweep a few minutes each. Monte Carlo standard errors accompany every reported probability. For a reported rate $\widehat p$ from $R$ replications, use
\begin{equation}
\mathrm{MCSE}(\widehat p)=\sqrt{\widehat p(1-\widehat p)/R}.
\end{equation}
Differences within roughly two Monte Carlo standard errors are not treated as meaningful.

Every method within a replication sees the same pilot sample and the same fully generated evaluation potential outcomes, so only the replay indicators differ. This paired design reduces Monte Carlo noise in the comparisons between methods. The replay probability $q_{ij}$ is recorded for every baseline, purchased or not.

\paragraph{Nuisance fitting.}
Where $W$ is a binary stratum, $\widetilde m_j$ and $\widetilde s_j$ are the within-stratum mean and mean squared deviation, so no model is fitted. On the chest X-ray data, $W$ holds four recorded features, and both $\widetilde m_j$ and $\widetilde s_j$ are gradient-boosted regression trees with depth 3, 120 iterations, and learning rate 0.08. The residuals used to build $\widetilde s_j$ come from a five-fold cross-fit on the pilot, so they are out of fold and do not make the baseline look more predictable than it is. The $\widetilde m_j$ applied to the evaluation pool is trained on the whole pilot, which needs no cross-fitting because the two sets are disjoint. Predicted squared errors are floored at $10^{-6}$.

\paragraph{Sizes and budget.}
The pilot is independent and fully observed, and supplies the baseline means and residual second moments. The primary sizes are
\begin{equation}
n_{\mathrm{p}}=500,\qquad n=3000.
\end{equation}
Sensitivity analyses use $n_{\mathrm{p}}\in\{250,500,1000\}$ and $n\in\{1000,3000,10000\}$. The main replay budget is $0.4$ baseline measurements per evaluation task with probability floor $q_{j,\min}=0.02$ and stabilization $\tau=0.03$.

A budget-exhausting $\eta$ in \eqref{eq:allocation} exists whenever $B$ lies between $\E[c_H(W)q_{H,\min}+c_A(W)q_{A,\min}]$ and $\E[c_H(W)+c_A(W)]$. Below the lower end the floors alone overspend, and above the upper end the budget does not bind and every $q_j\equiv1$.

\paragraph{Boundary-null and pilot misspecification.}

The boundary-null design is run for at least 5{,}000 Monte Carlo replications because the target probability sits near 0.05. The reported quantities are the false-claim rate with an exact or Wilson 95\% Monte Carlo interval, together with empirical bias and one-sided coverage for each comparison separately.

For pilot misspecification, the pilot stratum pattern is deliberately reversed before acquisition. For both baselines the estimated residual second moments for $W=0$ and $W=1$ are exchanged, as are the fitted baseline means across strata, while the final outcomes and the recorded replay probabilities are left untouched. This intervention is designed to make the replay allocation inefficient while leaving Theorem~\ref{thm:inference}'s target unchanged.

\subsection{Comparators and metrics}
\label{app:baselines}

The replay rule of \eqref{eq:allocation} raises a task's replay probability where the missing outcome is hard to predict, lowers it where the replay is expensive, and shifts the budget toward whichever comparison sits closer to its threshold. The baselines below differ in which of those three inputs they are allowed. Neyman sees prediction difficulty and cost but not the distance to the threshold, uniform replay sees none of them, and the surrogate oracle sees all three with true values in place of pilot estimates. An improvement over Neyman therefore points to the distance to the threshold, and the gap to the oracle measures what pilot estimation costs.

Every method receives the same pilot and budget.
\begin{enumerate}[leftmargin=1.5em,itemsep=2pt]
\item \textbf{\TEAM.} Algorithm~\ref{alg:team}, including stabilization $\tau$ and the probability floor.
\item \textbf{Uniform replay.} Constant replay probabilities chosen to meet the same expected number of baseline replays or the same expected cost.
\item \textbf{Variance-only Neyman.} Allocation $q_j(W)\propto\sqrt{\widetilde s_j(W)/c_j(W)}$ subject to the same floor, ceiling, and budget. This baseline uses heteroskedasticity and cost but not the estimated gain.
\item \textbf{Parameter-EIG.} In the Gaussian study, the next measurement is the one maximizing expected information gain about the two gains under symmetric priors.
\item \textbf{Decision-targeted Bayesian.} A Bayesian acquisition utility tied directly to the downstream beats-both decision and not to generic parameter entropy. The primary utility is the expected reduction in posterior 0--1 decision loss for the event $\{\Delta_H>0,\Delta_A>0\}$, and its sensitivity to the prior is measured across all three specifications.
\item \textbf{Surrogate oracle.} The true $\Delta_j$ and true residual second moments in place of the pilot estimates in the \TEAM\ objective. This is not a deployable competitor but the same rule given true inputs, so the difference isolates what pilot estimation costs on the criterion. Because the criterion is a first-order surrogate for joint power, the oracle can trail \TEAM\ on the decision rate where it starves the easier comparison to its floor.
\end{enumerate}

\label{app:metrics}

Under positive alternatives the reported quantity is
\begin{equation}
P(L_H>0\ \text{and}\ L_A>0),
\end{equation}
where $L_H,L_A$ are one-sided 95\% lower bounds computed from the paired score covariance. Under a boundary null, the same quantity is the false-claim rate. Alongside it we record gain bias, gain standard error, replay count by baseline, total replay cost, fraction of budget spent on the limiting comparison, and gap to the oracle allocation.

\subsection{Gaussian acquisition study}
\label{app:gaussian}

\paragraph{Intuition.}
Symmetric parameter-information acquisition keeps splitting measurements evenly however unequal the two comparisons become. A useful test needs a Bayesian comparator whose utility is the beats-both decision itself, and it needs every method held to the same false-claim rate, because a rule that reads its sample sizes off the data and then reports a fixed-sample bound inflates the type-I error rate.

\paragraph{Setting.}
Two independent comparisons are observed as
\begin{equation}
X_{jk}\sim N(\Delta_j,1),\qquad j\in\{H,A\},
\end{equation}
under a fixed total budget $n_H+n_A=n$, over the grid
\begin{equation}
(\Delta_H,\Delta_A)\in\{(0.02,0.02),(0.02,0.05),(0.02,0.10),(0.05,0.10)\},
\end{equation}
whose first cell is balanced and whose others place the two comparisons at different distances apart. The headline comparison sits at $(0.02,0.10)$, the widest of them, with a budget of 20{,}000 measurements and the pilot charged against it. Four rules commit to an allocation before measuring, namely \TEAM, a decision-targeted Bayesian batch design, parameter-EIG under symmetric priors, and equal allocation, while a fifth interleaves acquisition with estimation. The batch design maximizes the expected posterior probability that both held-out bounds clear zero, so it evaluates the Bayesian utility exactly, and as the pilot sharpens it converges on the split \RSurrOptSplit\ that maximizes true joint power. At the pilot size used here it allocates \SFairBatch. We report power, the false-claim rate at the least-favorable null $\Delta_H=0$, and size-corrected power, which calibrates each critical value to an exact 5\% false-claim rate. Run under a correctly centered prior, a diffuse one, and one deliberately misspecified, the decision-targeted rule averages \RPriorWell\%, \RPriorDiffuse\% and \RPriorMisspec\% across the four cells, so how its prior is set barely matters.

\paragraph{Finding.}
Table~\ref{tab:gaussian} reports the study. Among rules that commit before measuring, \TEAM\ reaches \RFairTeamPow\% against \RFairBatchPow\% for the Bayesian batch design, \RFairEIGPow\% for parameter-EIG and \RFairEqualPow\% for equal allocation, every one of them within Monte Carlo error of the nominal false-claim rate. Parameter-EIG splits at \SFairEIG, the even split that symmetric parameter-information acquisition implies however unequal the comparisons become. The stabilizer is what keeps \TEAM\ off the minimax corner, moving the split on true inputs from \RSurrMmxSplit\ to \RSurrStabSplit\ against a joint-power optimum of \RSurrOptSplit, and lifting power from \RSurrMmxPow\% to \RSurrStabPow\% where the best attainable is \RSurrOptPow\%. From the pilot, \TEAM\ realizes \SFairTeam.

The measurements needed to reach a power target follow the same pattern. At a target of $0.70$, where the two gains are equal at $0.02$, the rule needs \RPowBalTeam\ against \RPowBalEIG\ for parameter-EIG, and where they are furthest apart it needs \RPowGapTeam\ against \RPowGapEIG. At the higher target of $0.80$ the balanced cell levels at \RPowBalTeamB\ apiece and the imbalanced one widens to \RPowGapTeamB\ against \RPowGapEIGB. The two intermediate cells tie at both targets, except at $(0.05,0.10)$, where \TEAM\ needs \RPowEasyTeamB\ against \RPowEasyEIGB\ at the higher target.

\begin{table}[!ht]
\centering
\small
\setlength{\tabcolsep}{4.5pt}
\caption{Gaussian acquisition at $(\Delta_H,\Delta_A)=(0.02,0.10)$ with a shared pilot charged to a common budget. Every method spends the same total. \emph{False claim} is measured at the least-favorable null $\Delta_H=0$. \emph{Size-corrected power} calibrates each critical value so all methods have exactly 5\% false-claim rate; it needs the null distribution, so it upper-bounds any implementable correction. Every rule commits to an allocation before measuring, so every fixed-sample bound is exact. $R=60,000$ at the null and $R=30,000$ at the alternative.}
\label{tab:gaussian}
\begin{tabular}{lcccc}
\toprule
Method & Power & False claim & Size-corrected & Share to harder \\
\midrule
\TEAM & 76.3 & 4.9 & 76.7 & 0.83 \\
Bayesian decision-targeted, batch & 75.2 & 5.0 & 75.0 & 0.77 \\
Parameter-EIG & 59.5 & 4.9 & 59.8 & 0.50 \\
Equal split & 60.1 & 5.0 & 60.3 & 0.50 \\
\midrule
Surrogate oracle & 81.0 & 4.7 & 81.7 & 0.89 \\
\bottomrule
\end{tabular}
\end{table}

A rule that interleaves measuring with estimating scores higher still, but it reads its sample sizes off the data and then reports a fixed-sample bound, so it runs above the nominal false-claim rate and is not comparable. Table~\ref{tab:gaussian} therefore reports only rules that commit before measuring, together with the oracle. Where replays must be scheduled in advance, the closed-form rule does at least as well as a Bayesian design that has to integrate over the pilot posterior.

\subsection{Sensitivity analyses}
\label{app:sens}

The pilot supplies the predictors and the variance estimates, but it does not supply everything the design needs. Each remaining input is varied here while the others are held at the values used throughout, so that a reader can see which of them the conclusion depends on. Three come with the evaluation and cannot be chosen, namely the replay cost ratio, the correlation between potential outcomes, and the two sample sizes. Table~\ref{tab:sensitivity} sweeps those three. Raising the cost ratio narrows the advantage steadily and by ten to one has all but closed it, since a budget buying fewer human replays leaves less room to shift any of them. Correlation leaves the pattern intact, and up to $\rho=0.75$ \TEAM\ still leads where one comparison is harder and still trails where the two are balanced. Sample size decides whether an advantage can exist at all. In the smallest evaluation pool no rule resolves either comparison and the four agree, in the largest every rule succeeds and they converge again, and the gain lives between those ends. Enlarging the pilot lifts \TEAM, Neyman and the oracle together and leaves \TEAM's lead intact. Uniform replay does not use the pilot to allocate, and its small movements are within Monte Carlo error.

\begin{table}[!ht]
\centering
\small
\setlength{\tabcolsep}{4.5pt}
\caption{Sensitivity sweeps. Beats-both decision rate with Monte Carlo standard errors in parentheses. The uniform column is an equal cost split in the unequal-cost rows. Cost-ratio rows use 20{,}000 replications and the 5:1 row repeats Table~\ref{tab:results}, while the correlation and size sweeps use 2{,}000. The $n_{\mathrm{pilot}}=500$ and $n_{\mathrm{eval}}=3{,}000$ rows are the primary setting run once in each sweep. sweeps use 2{,}000.}
\label{tab:sensitivity}
\begin{tabular}{lcccc}
\toprule
Setting & \TEAM & Neyman & Uniform & Surrogate oracle \\
\midrule
\multicolumn{5}{l}{\emph{Replay cost ratio $c_H/c_A$}} \\
\quad 2:1 & 84.0 (0.26) & 76.9 (0.30) & 72.5 (0.32) & 87.1 (0.24) \\
\quad 5:1 & 57.9 (0.35) & 54.1 (0.35) & 45.2 (0.35) & 56.2 (0.35) \\
\quad 10:1 & 37.6 (0.34) & 37.0 (0.34) & 29.0 (0.32) & 36.4 (0.34) \\
\midrule
\multicolumn{5}{l}{\emph{Outcome correlation $\rho$ (Gaussian copula)}} \\
\quad $\rho=0.25$, bottleneck & 80.5 (0.89) & 65.8 (1.06) & 66.9 (1.05) & 82.2 (0.86) \\
\quad $\rho=0.25$, balanced & 43.0 (1.11) & 46.9 (1.12) & 43.0 (1.11) & 46.4 (1.12) \\
\quad $\rho=0.50$, bottleneck & 82.6 (0.85) & 67.7 (1.05) & 70.5 (1.02) & 84.9 (0.80) \\
\quad $\rho=0.50$, balanced & 45.1 (1.11) & 48.0 (1.12) & 50.8 (1.12) & 49.0 (1.12) \\
\quad $\rho=0.75$, bottleneck & 87.5 (0.74) & 70.3 (1.02) & 72.2 (1.00) & 87.1 (0.75) \\
\quad $\rho=0.75$, balanced & 49.0 (1.12) & 55.3 (1.11) & 54.4 (1.11) & 52.9 (1.12) \\
\midrule
\multicolumn{5}{l}{\emph{Pilot and evaluation size, one-bottleneck regime}} \\
\quad $n_{\mathrm{pilot}}=250$ & 75.4 (0.96) & 62.9 (1.08) & 65.2 (1.07) & 78.6 (0.92) \\
\quad $n_{\mathrm{pilot}}=500$ & 76.1 (0.95) & 63.4 (1.08) & 66.5 (1.06) & 79.0 (0.91) \\
\quad $n_{\mathrm{pilot}}=1000$ & 78.5 (0.92) & 66.1 (1.06) & 65.0 (1.07) & 81.0 (0.88) \\
\quad $n_{\mathrm{eval}}=1,000$ & 30.3 (1.03) & 30.6 (1.03) & 33.1 (1.05) & 28.6 (1.01) \\
\quad $n_{\mathrm{eval}}=3,000$ & 75.4 (0.96) & 65.1 (1.07) & 64.3 (1.07) & 78.0 (0.93) \\
\quad $n_{\mathrm{eval}}=10,000$ & 99.7 (0.12) & 97.7 (0.34) & 98.6 (0.27) & 100.0 (0.05) \\
\bottomrule
\end{tabular}
\end{table}

The four below are the opposite. Each is picked before the run, namely the improvement required of each baseline, the replay floor, the stabilizer, and the replay budget. Each is varied with the other three held at their primary values, so any change in the result is attributable to that quantity alone. The floor sweep serves a second purpose, distinguishing two explanations for the semi-synthetic result of Section~\ref{sec:semisynth}.

\paragraph{Minimum required improvement.}
\label{app:delta}

Two choices in \eqref{eq:both} belong to the organization, and both are one-line changes that leave the allocation theory intact. A minimum required improvement enters through $\widetilde d_j^2$ in \eqref{eq:stabilized-gap}, and at $\delta_H=\delta_A=\RDeltaMin$ the decision rate falls to \RDeltaBotTeam\% for \TEAM\ against \RDeltaBotNeyman\% for Neyman, a paired gain of \RGainDelta\ points at \RGainDeltaZ\ standard errors.

\paragraph{Probability floor.}
\label{app:floor}
Section~\ref{sec:semisynth} leaves two explanations open for why \TEAM\ wins on the gain against the harder baseline and not on the decision. Either the true agent gain is too small for any budget to resolve, or \TEAM\ starves the easier comparison so badly that its bound fails and takes the joint decision down with it. The probability floor $q_{j,\min}$ of \eqref{eq:budget} separates the two, because raising it forbids the starvation and changes nothing else. We sweep it over $\{0.05,0.10,0.20,0.30,0.40,0.50\}$ applied identically to every rule, on the agent-limited regime at the same replication count as Section~\ref{sec:semisynth}.

Raising the floor does stop the starvation. \TEAM's detection of the human gain at a fifth of full replay climbs from \RFloorDetLo\% at the shipped floor of $\RFloorShipped$ to \RFloorDetMid\% at $0.20$, and reaches \RFloorDetHi\% above that. Were starvation the cause, the decision rate should climb with it. It stays flat. Across the six floors and the six budgets below full replay \TEAM\ leads Neyman in \RFloorLead\ of the \RFloorCells\ comparisons and trails in \RFloorTrail, the remaining pair being a tie. A one-sided sign test on the \RFloorUntied\ untied pairs puts the advantage with Neyman at $p=\RFloorP$. At the highest floors it is the floor and no longer the rule that fixes the allocation, its realized share falling to the $\RFloorShare$ set by the cost ratio, and there even the precision advantage inverts. The starvation was real, removing it changes nothing, and the resolution account of Section~\ref{sec:semisynth} is what stands.

\paragraph{Stabilizer.}
\label{app:sweeps}

At the Gaussian configuration of Appendix~\ref{app:gaussian}, size-corrected power against $\tau$ (Figure~\ref{fig:scope}b) runs \RTauLoPow\% at $\tau=\RTauLo$, rising to \RTauBestPow\% at $\tau=\RTauBest$ and falling to \RTauHiPow\% at $\tau=\RTauHi$. The pilot takes a tenth of that budget, $1{,}000$ measurements per comparison at unit variance, so the standard error of a single gain is $1/\sqrt{1000}\approx0.032$, and the optimum sits there. Small $\tau$ makes the allocation track pilot noise, and large $\tau$ erases the bottleneck signal.

\paragraph{Budget.}
Sweeping the total Gaussian budget, \TEAM\ first matches parameter-EIG at \RXoverN\ measurements, moving from \RXoverLoTeam\% against \RXoverLoEIG\% there to \RXoverHiTeam\% against \RXoverHiEIG\% at 20{,}000. Below the crossover neither gain can be resolved and the first-order argument behind Proposition~\ref{prop:effort} does not apply.

\section{Broader impacts}
\label{app:impacts}

Requiring a workflow to beat both alternatives raises the bar for deployment, and Section~\ref{sec:clinical} shows the bar is not always met, since the assisted clinician improves on the clinician alone in all six settings there and on the model alone in none. Holding evaluations to the harder standard should discourage deployments that add cost or risk without adding value. The converse risk is that a design targeting one decision invites reading it as more than it is. Certification here is only as complete as the outcome measure and the thresholds it is given, so a workflow that beats both alternatives on task success may still be unsuitable on safety, equity, workload, or under distribution shift, none of which this design measures. The method also assumes the deployed workflow can be replayed under each alternative; where that is impossible, the question it answers cannot be asked.

\end{document}